\documentclass[letterpaper, 10 pt, conference]{ieeeconf}
\IEEEoverridecommandlockouts  
\usepackage{xspace}
\usepackage[dvipsnames]{xcolor}
\usepackage{graphicx}
\usepackage{amsmath}
\usepackage{amssymb}
\usepackage{booktabs}
\usepackage{cite}
\usepackage{etoolbox}
\usepackage[hyphens]{url}
\let\labelindent\relax 
\usepackage[shortlabels,inline]{enumitem}
\setlist[itemize]{noitemsep,leftmargin=*,topsep=0em}
\setlist[enumerate]{noitemsep,leftmargin=*,topsep=0em}
\makeatletter
\DeclareRobustCommand\onedot{\futurelet\@let@token\@onedot}
\def\@onedot{\ifx\@let@token.\else.\null\fi\xspace}

\makeatother

\newcommand{\tabnotes}[1]{\par\smallskip\parbox{\linewidth}{\footnotesize\leftskip=0pt\rightskip=0pt\parfillskip=0pt plus 1fil #1}} 
\usepackage{capt-of} 

\usepackage{multirow}
\usepackage{makecell}
\usepackage{bm}
\usepackage{amsthm}

\newcommand{\ours}{NibbleFlow\xspace}
\newcommand{\sem}{SEM\xspace}           
\newcommand{\pow}[1]{2^{#1}}

\providecolor{bestF32}{HTML}{0B7F87}
\providecolor{bestQ}{HTML}{FF7F0E}
\providecolor{semgreen}{HTML}{006400}
\AtBeginEnvironment{tabular}{\renewcommand{\sem}{\textcolor{semgreen}{SEM}\xspace}}
\AtBeginEnvironment{tabular*}{\renewcommand{\sem}{\textcolor{semgreen}{SEM}\xspace}}

\makeatletter
\renewcommand\paragraph{\@startsection{paragraph}{4}{\parindent}{1.5ex plus .5ex minus .2ex}{0ex}{\normalfont\normalsize\itshape}}
\makeatother

\definecolor{linkblue}{rgb}{0.21,0.49,0.74}
\usepackage[breaklinks,colorlinks,allcolors=linkblue,pdfauthor={Xiao'ao Song, Chahat Deep Singh},pdftitle={NibbleFlow: Sign-Exponent-Mantissa Refinement for 4-Bit Optical Flow}]{hyperref}
\usepackage[capitalize]{cleveref}
\crefname{section}{Sec.}{Secs.}
\Crefname{section}{Section}{Sections}
\Crefname{table}{Table}{Tables}
\crefname{table}{Table}{Tables}
\newif\ifsupp
\suppfalse

\title{\LARGE \bf NibbleFlow: Sign-Exponent-Mantissa Refinement for 4-Bit Optical Flow}

\author{Xiao'ao Song and Chahat Deep Singh%
\thanks{The authors are with the Perception, Robotics, AI and Sensing (PRAISe) Lab, University of Colorado Boulder, USA.
Email: \{xiaoao.song, chahat.singh\}@colorado.edu.}%
\\[2pt]
{\normalsize\url{https://www.praisecu.com/research/NibbleFlow}}%
}

\AtBeginDocument{\raggedbottom}
\begin{document}
\IEEEaftertitletext{
\begin{center}
\includegraphics[width=\textwidth]{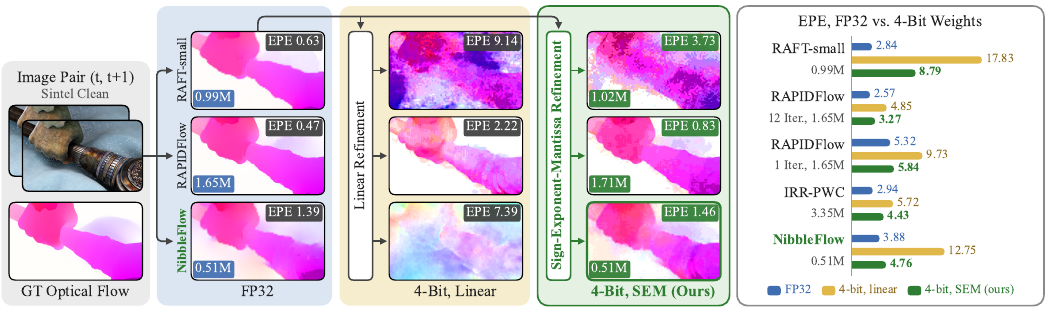}
\captionof{figure}{\textcolor{semgreen}{S}ign-\textcolor{semgreen}{E}xponent-\textcolor{semgreen}{M}antissa refinement for 4-bit optical flow.
Each network is shown in FP32 and with 4-bit weights, with its linear output layer or with \textbf{\textcolor{semgreen}{SEM}}.
The 4-bit results use post-training quantization to 4-bit weights and 8-bit activations, except \ours with \sem, which runs as INT4 on the Hailo-8 (\cref{tab:main}).
Left, one Sintel clean pair (\texttt{ambush\_7}) with the EPE and parameter count in each panel. The FP32 panels of RAFT-small and RAPIDFlow use \sem.
Right, Sintel clean EPE (\cref{tab:main}).
\sem significantly reduces the 4-bit error of every network.}
\label{fig:teaser}
\end{center}
}
\maketitle
\thispagestyle{plain}
\pagestyle{plain}
\begin{abstract}

Accurate optical flow relies on iterative refinement, yet the edge accelerators that small robots and other devices on batteries carry execute only static integer graphs and often leave just 4 bits per weight.
Under this constraint, iterative networks lose much of their accuracy, and the networks designed for these chips are far less accurate to begin with.
We trace this loss to a single layer, the linear output of the update block.
Refinement needs steps whose sizes differ by more than a factor of one hundred, but a linear output represents them like a fixed point number.
The rounding error of its weights therefore adds the same error to every update, however small, and this error accumulates over the iterations.
We introduce \textbf{\textcolor{semgreen}{SEM}}, an output layer that instead predicts each flow update as a floating point number with a \textbf{\textcolor{semgreen}{S}}ign, an \textbf{\textcolor{semgreen}{E}}xponent and a \textbf{\textcolor{semgreen}{M}}antissa.
A rounding error then perturbs each update only in proportion to its size, while every output remains bounded and easy to quantize.
Replacing only the output layer of existing networks such as RAFT-small and RAPIDFlow preserves their floating point accuracy and removes up to 86\% of the error that 4-bit weights add.
Built around SEM, NibbleFlow has only 0.51M parameters and runs on a Hailo-8 with most of its weights in 4 bits.
It loses less than one pixel of EPE to its floating point version and cuts the Sintel error of the best deployable baseline by up to 45\%, with $5.4\times$ fewer parameters.

\end{abstract}
\section{Introduction}
\label{sec:intro}

Optical flow underpins visual odometry~\cite{pandey2021leveraging}, structure from motion~\cite{schonberger2016structure}, tracking~\cite{karpuzov2024object} and navigation~\cite{chao2014survey}, and its accuracy today is driven by iterative refinement~\cite{teed2020raft,sun2018pwc}, in which an update block repeatedly corrects the current estimate.
These methods run in floating point on desktop GPUs, whereas small robots and other battery-powered devices rely on edge AI accelerators such as the Hailo-8 and the Coral Edge TPU, which deliver several TOPS within a few watts~\cite{edgeflownet2024,krispin2024real}.
This efficiency comes at a price, since these accelerators execute only a static graph with integer weights and keep all weights in the memory on the chip, which often leaves only 4 bits per weight.
Under such quantization, iterative flow networks lose much of their accuracy, while the networks that were designed for these chips are far less accurate to begin with.
As a result, accurate optical flow has remained out of reach on the very hardware that needs it.

Two obstacles stand between iterative refinement and such hardware.
The first is (i) the warp.
At every iteration, refinement resamples the features of the second frame at positions shifted by the current flow~\cite{song2026tinyflow}.
Because the memory addresses of this operation depend on the data, a static graph cannot express it.
For this reason, the networks that deploy on these accelerators, such as NanoFlowNet~\cite{bouwmeester2023nanoflownet} and EdgeFlowNet~\cite{edgeflownet2024}, give up iteration entirely and pay for it in accuracy.
TinyFlow~\cite{song2026tinyflow} recently showed that iteration can be kept without a warp.

Removing the warp, however, is not enough.
A second obstacle remains, (ii) \emph{how the update block represents a step}, and it is the subject of this paper.
Iterative refinement asks a single output layer for steps of very different sizes, because the first iterations must cover most of the motion while the last ones only polish it.
To quantify this demand, we measure step sizes in octaves, where one octave is a factor of two.
At the first step of our coarsest level alone, the needed displacement already spans seven octaves, from $\pow{-6}$ to $\pow{1}$ pixels, and every octave holds a similar share of the flow components.
A standard linear output head is poorly suited to such a range, because it represents a step the way a fixed point number represents a value.
Its error is absolute, so this error decides the smallest step the head can take accurately, while the largest value of the last layer decides the largest one.
Our measurements confirm this picture.
The linear update covers only 5.6 octaves at the first step (\cref{sec:exp_steps}).

Motivated by this observation, we replace the linear step with a floating point one, the \textbf{\textcolor{semgreen}{S}}ign, \textbf{\textcolor{semgreen}{E}}xponent and \textbf{\textcolor{semgreen}{M}}antissa (\textbf{\textcolor{semgreen}{SEM}}) update.
For each flow component, the update block predicts a sign, a distribution over exponents that are powers of two and a mantissa, and the step is their product.
Every predicted quantity is bounded, so the head remains easy to quantize, but the network now chooses the scale of each step at each pixel.
An error in the head therefore becomes an error relative to the step, and the rounding error of 4-bit weights no longer adds a fixed error to every update.
As a consequence, \sem covers all seven octaves.
Because only the output layer changes, \sem can replace the linear layer of any network that refines its flow by increments.
In RAFT-small~\cite{teed2020raft} and RAPIDFlow~\cite{morimitsu2024rapidflow}, it preserves the floating point accuracy and removes up to 86\% of the error that the conversion from FP32 to 4-bit weights adds (\cref{tab:main}).
Finally, we build \ours around this update to show what it enables on a real accelerator.
\ours refines flow without a warp, has only 0.51M parameters and runs on a Hailo-8 with most of its weights in 4 bits.
It loses less than one pixel of EPE to its FP32 version on Sintel and is the most accurate deployable network in our comparison (\cref{sec:exp_main}).

In summary, our contributions are as follows.
\begin{itemize}[leftmargin=*,itemsep=1pt,topsep=2pt]
  \item We introduce \textbf{\textcolor{semgreen}{SEM}}, a floating point output layer that predicts each flow update as a \textbf{\textcolor{semgreen}{S}}ign, an \textbf{\textcolor{semgreen}{E}}xponent and a \textbf{\textcolor{semgreen}{M}}antissa.
  Where a standard linear output has a fixed dynamic range, \sem covers the full range of steps that refinement needs.
  \item We show that \sem makes existing iterative networks robust to 4-bit weights.
  It replaces their linear output layer without tuning, preserves their floating point accuracy and removes up to 86\% of the error that 4-bit weights add.
  \item We propose \ours, a flow network without a warp that is built around \sem and runs on a Hailo-8 with most of its weights in 4 bits.
  On this real accelerator, \ours loses less than one pixel of EPE to its FP32 version and is the most accurate deployable network in our comparison.
\end{itemize}
\section{Related Work}
\label{sec:related}

\paragraph{Iterative optical flow}
FlowNet~\cite{dosovitskiy2015flownet} established that optical flow can be learned from end to end.
PWC-Net~\cite{sun2018pwc} and LiteFlowNet~\cite{hui2018liteflownet} then made feature warping and local cost volumes the standard tools of coarse to fine estimation.
RAFT~\cite{teed2020raft} changed this picture by replacing the pyramid with a recurrent update operator that looks up a correlation volume over all pairs at the current flow in every iteration, a line that SEA-RAFT~\cite{wang2024searaft} and GMFlow~\cite{xu2022gmflow} refine further.
Since then, much effort has gone into making refinement cheaper.
IRR-PWC~\cite{hur2019irr} shares one decoder across iterations, while RAPIDFlow~\cite{morimitsu2024rapidflow}, FastFlowNet~\cite{kong2021fastflownet} and NeuFlow~\cite{zhang2025neuflow} reduce its cost for embedded GPUs.
Despite their differences, the iterative methods share two design choices.
They align the frames by resampling at the current flow, and they regress each update as a linear quantity.
The first choice prevents deployment on static accelerators, and the second is the subject of this paper.

\paragraph{Optical flow on edge accelerators}
Only a few flow networks run on integer accelerators.
NanoFlowNet~\cite{bouwmeester2023nanoflownet} runs on a GAP8 microcontroller and EdgeFlowNet~\cite{edgeflownet2024} on a Coral Edge TPU, but both fit the accelerator by predicting flow in a single forward pass and thus give up refinement.
TinyFlow~\cite{song2026tinyflow} showed that refinement can be kept.
It correlates the frames over fixed offsets with convolutions of constant kernels and feeds the current flow to the update block, so that the network compiles as one static INT8 graph.
We adopt its mechanism for removing the warp.
However, we show that its linear update limits both the accuracy and the tolerance of 4-bit weights, and we replace it with \sem.

\paragraph{Output parameterization}
A long line of work replaces the regression of a continuous target with classification over bins.
Depth networks classify each pixel into depth bins that are ordered~\cite{fu2018dorn} or adapted to each image~\cite{bhat2021adabins}.
Stereo and flow methods read disparity or displacement out of a cost volume with a soft argmin~\cite{kendall2017gcnet,xu2022gmflow}, and detection heads predict box edges as distributions over bins~\cite{li2020gfl}.
The HL-Gauss loss trains generic regressors in the same way~\cite{farebrother2024stop}.
All of these methods discretize the \emph{value} itself.
Closer to our work, a few methods split a predicted number into an exponent and a mantissa, for ground sampling distance~\cite{lee2019gsd}, for numbers as transformer tokens~\cite{charton2022linear} and for rewards in signed log space~\cite{hafner2025dreamerv3}.
We bring this split to iterative refinement, where it matters most.
\sem applies it to every flow update and classifies only the \emph{order of magnitude} of the step, which is precisely what changes from one iteration to the next.

\paragraph{Quantization}
Integer inference with fake quantization during training~\cite{jacob2018quantization,nagel2021white} is the standard route to accelerators.
To gain dynamic range, floating point and logarithmic formats such as FP8~\cite{micikevicius2022fp8} and log quantization~\cite{miyashita2016logarithmic} give up uniform resolution in the \emph{weights and activations}.
We make the same trade in the \emph{predicted quantity} instead, which remains easy to quantize because each of its parts is bounded.

\section{Method}
\label{sec:method}

\begin{figure*}[t]
  \centering
  \includegraphics[width=\linewidth]{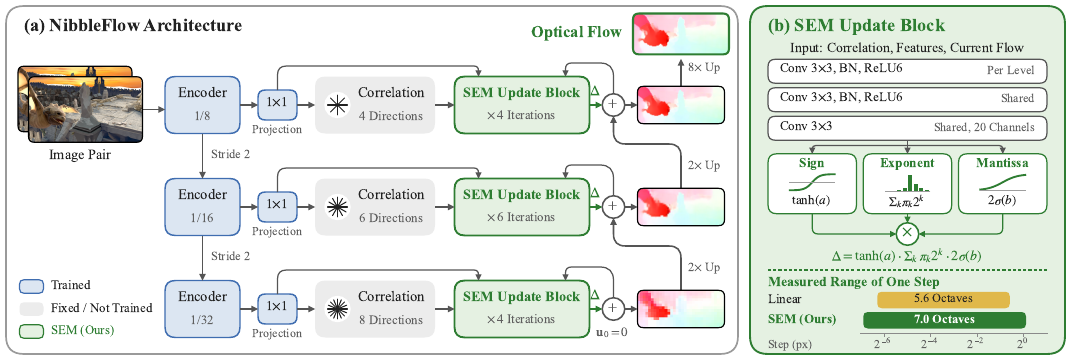}
  \caption{\ours architecture.
  (a)~Features at $1/8$, $1/16$ and $1/32$ are correlated along fixed directions, so nothing is warped, and the flow is refined coarse to fine from zero.
  (b)~The \sem update block predicts each step as the product of a sign, an exponent and a mantissa (\cref{eq:em}). The range of one step is measured in \cref{sec:exp_steps}.}
  \label{fig:arch}
\end{figure*}

The key idea of our method is to represent each refinement step as a floating point number instead of a fixed point one.
We first recall the refinement scheme that we build on (\cref{sec:warpfree}).
We then analyze the limitations of the linear update (\cref{sec:linear}) and introduce \sem (\cref{sec:em}).
Finally, we describe the network and its training (\cref{sec:arch,sec:training}).

\subsection{Preliminaries}
\label{sec:warpfree}
We build on the refinement scheme of~\cite{song2026tinyflow}, which compiles to a static graph because it never warps.
Given two frames $I_1$ and $I_2$, a shared encoder produces features at strides $s \in \{8, 16, 32\}$, and a $1{\times}1$ projection normalizes them to $\hat f^s_1$ and $\hat f^s_2$ with operations that fuse into one integer convolution.
Instead of warping the second frame, each level correlates the two feature maps over a fixed set of offsets $\mathcal{D}_s$, that is, $C^s(\mathbf{x}, \boldsymbol{\delta}) = \langle \hat f^s_1(\mathbf{x}),\, \hat f^s_2(\mathbf{x}+\boldsymbol{\delta}) \rangle$ for $\boldsymbol{\delta}\in\mathcal{D}_s$.
Because every offset is constant, this volume is built from convolutions with fixed kernels and an elementwise product, and a window of radius $r$ reaches $\pm r s$ pixels of the input image.
Refinement at level $s$ starts from the upsampled estimate of the coarser level, $\mathbf{u}^s_0 = 2\,\mathrm{Up}(\mathbf{u}^{2s})$, or from zero flow at the coarsest level, and applies $T_s$ updates
\begin{equation}
  \mathbf{u}^s_{t+1} = \mathbf{u}^s_t + \Delta^s_t, \qquad \Delta^s_t = \mathcal{H}_s\big(C^s, \hat f^s_1, \mathbf{u}^s_t\big),
  \label{eq:refine}
\end{equation}
in pixels of level $s$.
The correlation is never recentered.
Instead, the current flow enters the update block $\mathcal{H}_s$ as two input channels, so the dependence on the data is carried in values rather than in memory addresses.
This scheme settles where the update block looks, but not how it expresses its answer.
The rest of this section addresses that question, namely how $\mathcal{H}_s$ should represent the step $\Delta$.

\subsection{Limitations of the Linear Update}
\label{sec:linear}
Iterative flow networks regress each update linearly.
Let $g$ denote the output of the last convolution of $\mathcal{H}_s$, with one channel per flow component.
The linear update is simply $\Delta = g$.
To understand what such an update can express, we compare it with the step that refinement needs, that is, the residual $\rho_t = u^{\star,s} - u^s_t$ between the ground truth at that level and the current flow.

These needed steps span many octaves, where one octave is a factor of two.
At the first step of our $1/32$ level, they range from $\pow{-6}$ to $\pow{1}$ pixels, and every octave between $\pow{-6}$ and $\pow{-1}$ alone holds between 9\% and 20\% of the flow values.
A linear update is poorly matched to this distribution.
To see why, suppose that the last layer outputs $g+\epsilon$ instead of $g$, where $|g|\le G$ and the error has a magnitude $|\epsilon| = e$ that does not depend on $g$.
Such an error arises from limited capacity and feature noise and, after deployment, mainly from the rounding of 4-bit weights (\cref{sec:exp_general}).
For a linear update $\Delta=g+\epsilon$, the relative error of a step is $|\epsilon|/|g|$.
A step is therefore accurate to a relative error $\eta$ only if $|g|\ge e/\eta$, so the accurate steps lie in $[e/\eta,\ G]$.
We call the span of this interval the \emph{dynamic range} of the update,
\begin{equation}
  \log_2\frac{G}{e/\eta} = \log_2\frac{\eta\, G}{e} \quad \text{octaves}.
  \label{eq:range}
\end{equation}
For instance, an error of $\pow{-3}$ pixels is eight times larger than a needed step of $\pow{-6}$ pixels, so this step is lost.

In other words, the error of a linear update is \emph{absolute}, so the window of accurate steps has a fixed width.
This is exactly the behavior of a fixed point number, whose resolution is uniform and whose range is fixed by its word length.
Our measurements on trained networks agree with this analysis.
In \cref{sec:exp_steps}, the measured dynamic range of the linear update is only 5.6 octaves.
A linear update therefore cannot cover the range that refinement needs.

\subsection{\texorpdfstring{\textcolor{semgreen}{S}ign-\textcolor{semgreen}{E}xponent-\textcolor{semgreen}{M}antissa (\textcolor{semgreen}{SEM}) Refinement}{Sign-Exponent-Mantissa (SEM) Refinement}}
\label{sec:em}
A floating point number escapes this limit because it stores the order of magnitude separately from the significant digits.
We let the update block do the same.
For each flow component, $\mathcal{H}_s$ predicts three quantities instead of one, namely a sign logit $a$, exponent logits $\boldsymbol{\ell}\in\mathbb{R}^K$ over the powers of two $\pow{k}$ with $k\in\{k_{\min},\dots,k_{\max}\}$, and a mantissa logit $b$.
The step is their product,
\begin{equation}
  \begin{aligned}
  \Delta &= \tanh(a)\cdot E\cdot 2\sigma(b),\\
  E &= \sum_{k=k_{\min}}^{k_{\max}} \pi_k\,\pow{k}, \quad \boldsymbol{\pi} = \mathrm{softmax}(\boldsymbol{\ell}).
  \end{aligned}
  \label{eq:em}
\end{equation}
Each factor has a distinct role.
The sign $\tanh(a)$ gives the direction of the step and can shrink it smoothly to zero, the exponent $E$ selects its order of magnitude, and the mantissa $2\sigma(b)$ refines it continuously.
The softmax interpolates between the exponents instead of selecting one, so that the step remains differentiable and $E$ can take any value between $\pow{k_{\min}}$ and $\pow{k_{\max}}$.

The benefit of this form becomes clear when we take the logarithm of \cref{eq:em}, which turns the product into the sum $\log_2|\Delta| = \log_2|\tanh a| + \log_2 E + \log_2 2\sigma(b)$.
The outputs of the network therefore set the logarithm of the step, so an error in them becomes a \emph{relative} error of the step and no longer an absolute one.
More precisely, an error $\epsilon$ in $a$ changes $\Delta$ by at most $|\epsilon|\,B$, where $B = E\cdot 2\sigma(b)$ is the scale that the network chose for that pixel.
As a result, a step of $\pow{-6}$ pixels and a step of $\pow{2}$ pixels are taken with the same relative precision.
In contrast to \cref{eq:range}, the dynamic range of \sem is thus set by the span of its exponents and no longer by the error $e$.

\paragraph{Choosing the exponents}
A natural concern is that the exponent range becomes a new parameter to tune.
This is not the case, because the exponents are independent of each other.
Adding a larger power of two leaves the smaller ones unchanged, and with them the small steps, as long as the softmax gives it little weight where it is not needed.
By contrast, a linear update can reach larger steps only by giving up its smallest ones (\cref{sec:linear}).
The largest exponent therefore only has to cover the largest step that the network can take, which is fixed by its search range, and exponents beyond it would simply go unused.
For \ours we use $k\in\{-6,\dots,1\}$, since the largest step, $2\cdot\pow{1}$, then equals the correlation radius of 4 feature pixels, which is the largest displacement that a level can see.
This gives $K=8$ exponents and $2\,(2+K) = 20$ output channels.
The existing networks of \cref{sec:exp_general} keep the same smallest exponent and extend the largest one to their own search range, up to $\pow{5}$ for optical flow (\cref{fig:ptq_qual}).

\paragraph{Suitability for integer hardware}
\sem is also well suited to integer arithmetic.
Every predicted quantity lies in a fixed interval, namely $\tanh(a)\in(-1,1)$, $\boldsymbol{\pi}\in[0,1]^K$ and $2\sigma(b)\in(0,2)$, so each can be quantized to 8 bits with a static scale.
Moreover, decoding the step requires only a softmax, a fixed $1{\times}1$ convolution with weights $\pow{k}$ and two elementwise products, all of which edge accelerators support.
Most importantly, the rounding of the weights and inputs of the output layer acts like the error $\epsilon$ of \cref{sec:linear}, and \sem turns it into an error relative to the step.
This is why replacing the linear output layer of RAFT-small and RAPIDFlow by \sem removes up to 86\% of the error that 4-bit weights add (\cref{sec:exp_general}).

\subsection{\ours Architecture}
\label{sec:arch}
We now build a network around \sem that an edge accelerator can run as a single static graph.
\Cref{fig:arch} shows the result, which follows the scheme of \cref{sec:warpfree} and adds three design choices.

\paragraph{Encoder}
The encoder is built from depthwise separable blocks (DW$3{\times}3$, \allowbreak BN, \allowbreak ReLU6, \allowbreak PW$1{\times}1$, \allowbreak BN, \allowbreak ReLU6), where DW and PW denote depthwise and pointwise convolutions and BN batch normalization, and from residual blocks with a ReLU6 after the addition.
It produces 64, 56 and 64 channels at $1/8$, $1/16$ and $1/32$.

\paragraph{Directional correlation}
Without a warp, a level sees only the motion that falls inside its correlation window, so large motion must be captured at a coarse level.
We therefore start the refinement at $1/32$, where a radius of 4 feature pixels reaches $\pm128$ pixels of the input image.
To keep the cost low, we do not correlate over a full window but only along a few directions through each pixel, namely eight directions $22.5^\circ$ apart at $1/32$, six directions $30^\circ$ apart at $1/16$ and four at $1/8$, each rasterized to the nearest pixel.
In this way, the correlation adds channels only where the grid is small.

\paragraph{Shared update block}
All three levels use a single \sem update block of width 96 with three $3{\times}3$ convolutions; the first two are each followed by BN and ReLU6, and the output convolution by neither.
Its hidden and output layers are shared across the levels, which keeps the network small.
Only the input layer is specific to a level, because the number of correlation channels differs, together with the batch normalization statistics.
Refinement then proceeds from coarse to fine with $T_{32}/T_{16}/T_8 = 6/6/3$ iterations during training and $4/6/4$ at inference, which gives the lowest EPE on the FlyingChairs training set among the settings with fewer iterations than training.
The final flow is the $1/8$ estimate after the last iteration, bilinearly upsampled to full resolution.

\subsection{Training}
\label{sec:training}
\setcounter{dbltopnumber}{1}
\begin{table*}[t]
\centering
\caption{Comparison with edge optical flow networks}
\label{tab:main}
\footnotesize
\setlength{\tabcolsep}{3.5pt}
\begin{tabular}{@{}l c c ccc ccc ccc ll@{}}
\toprule
 & & & \multicolumn{3}{c}{Clean} & \multicolumn{3}{c}{Final} & \multicolumn{3}{c}{Chairs} & \multicolumn{2}{c}{FPS} \\
\cmidrule(lr){4-6}\cmidrule(lr){7-9}\cmidrule(lr){10-12}\cmidrule(lr){13-14}
 & & & & \multicolumn{2}{c}{INT4} & & \multicolumn{2}{c}{INT4} & & \multicolumn{2}{c}{INT4} & & \\
\cmidrule(lr){5-6}\cmidrule(lr){8-9}\cmidrule(lr){11-12}
Method & Params (M) & GMACs & FP32 & Linear & \sem & FP32 & Linear & \sem & FP32 & Linear & \sem & RPi~5 & Hailo-8 \\
\midrule
\multicolumn{14}{@{}l}{\emph{Reference iterative models, not deployable}} \\
RAFT-small$^\S$~\cite{teed2020raft} & 0.99 & 211.78 & 2.84 & 17.83 & 8.79 & 4.06 & 19.70 & 10.46 & \underline{1.71} & 8.96 & 3.89 & 0.07 & -- \\
RAPIDFlow-it12$^\S$~\cite{morimitsu2024rapidflow} & 1.65 & 54.98 & \underline{2.57} & \underline{4.85} & \underline{3.27} & \underline{3.59} & \underline{5.83} & \underline{4.28} & \underline{1.71} & 3.55 & \underline{2.33} & 0.40 & -- \\
RAPIDFlow-it1$^\S$ & 1.65 & 5.85 & 5.32 & 9.73 & 5.84 & 6.07 & 10.25 & 6.57 & 4.21 & 8.69 & 4.83 & \underline{2.12} & -- \\
IRR-PWC$^{\S\ddagger}$~\cite{hur2019irr} & 3.35 & 90.93 & 2.94 & 5.72 & 4.43 & 4.18 & 6.72 & 5.66 & 1.83 & \underline{3.27} & 2.70 & 0.29 & -- \\
\midrule
\multicolumn{14}{@{}l}{\emph{Deployable on an integer edge accelerator}} \\
NanoFlowNet~\cite{bouwmeester2023nanoflownet} & \textbf{0.17} & \textbf{0.93} & 9.35 & 14.14 & -- & 9.83 & 14.31 & -- & 7.04 & 10.39 & -- & \textbf{11.24} & \textbf{120.55} \\
EdgeFlowNet~\cite{edgeflownet2024} & 2.75 & 19.98 & 7.41 & 8.65 & -- & 8.25 & 9.22 & -- & 3.56 & 5.50 & -- & 1.02 & 76.30 \\
\textbf{\ours{} (ours)} & 0.51 & 6.97 & \textbf{3.88} & 12.75 & \textbf{4.76} & \textbf{4.76} & 13.30 & \textbf{5.68} & \textbf{2.45} & 9.15 & \textbf{3.14} & 2.25 & 15.19 \\
\bottomrule
\end{tabular}
\tabnotes{Bold marks the best deployable result and underlining the best reference result. $^\S$Trained on FlyingChairs only and quantized in simulation. $^\ddagger$Two layers kept in 8 bits (\cref{sec:exp_setup}).}
\end{table*}
\begin{figure*}[t]
  \centering
  \includegraphics[width=\linewidth]{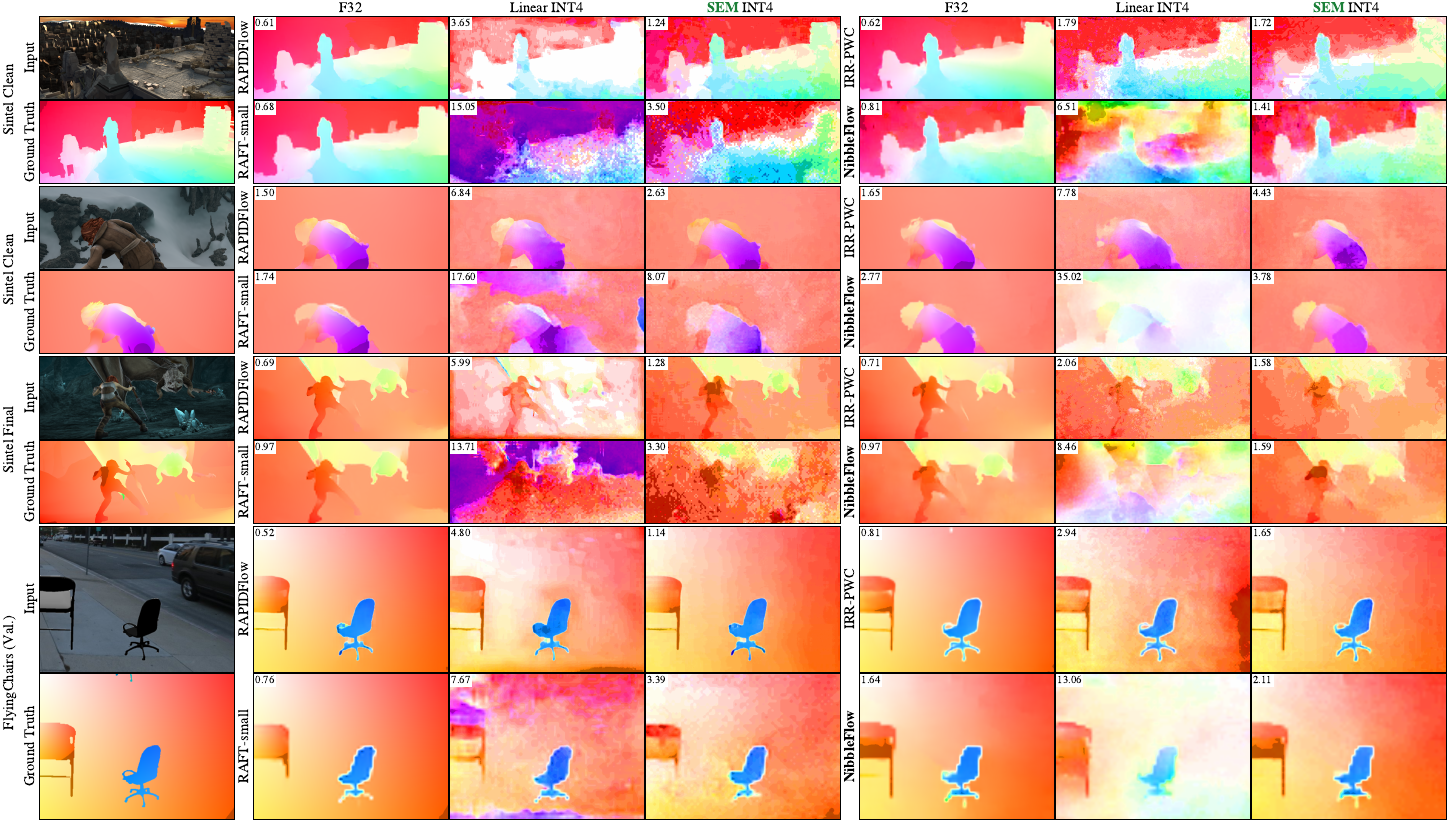}
  \caption{The linear output layer and \sem in four iterative flow networks.
  Each network is shown in FP32 and in INT4, with its linear output layer or with \sem, and the EPE of the pair is given in each panel.
  With the linear layer, every network loses much of the flow in INT4, whereas with \sem the flow stays close to FP32.}
  \label{fig:ptq_qual}
\end{figure*}
We supervise every iterate of every level against the ground truth downsampled to that level, $\mathbf{u}^{\star,s}$, using a Charbonnier penalty $\rho$ and the exponential weights of RAFT~\cite{teed2020raft},
\begin{equation}
  \mathcal{L} = \sum_{s} \sum_{t=1}^{T_s} \frac{\gamma^{T_s-t}}{\sum_j\gamma^{T_s-j}}\,\rho\big(\mathbf{u}^s_t - \mathbf{u}^{\star,s}\big) + w\,\mathcal{L}_{\mathrm{smooth}},
  \label{eq:loss}
\end{equation}
where $\rho(\mathbf{x}) = \sum_c (x_c^2+\epsilon^2)^{\alpha}$ over the two flow components with $\alpha=0.45$ and $\epsilon=10^{-3}$, $\gamma=0.8$, and $\mathcal{L}_{\mathrm{smooth}}$ is an edge-aware smoothness term on the $1/8$ flow with $w = 0.05$.
The weights are normalized at each level, so every level contributes equally whatever its number of iterations.
We first train on FlyingChairs~\cite{dosovitskiy2015flownet} and then fine-tune on a mixture of FlyingChairs (55\%) and the FlyingThings3D subset~\cite{mayer2016things3d} (45\%).
In addition, we initialize a level from a corrupted version of the ground truth, with smooth noise of up to 2 pixels of that level, so that it learns to correct imperfect inputs rather than rely on the quality of its initial flow.
On FlyingChairs this happens with probability 0.16, 0.08 and 0.04 at $1/32$, $1/16$ and $1/8$; during fine-tuning it is 0 at $1/8$, and in its last stage 0.1 and 0.05 at $1/32$ and $1/16$.
Finally, we quantize the fine-tuned network with the Hailo compiler and run its quantization-aware fine-tuning (QFT), which distills the floating point network into the quantized one.
QFT uses 1024 pairs of the same FlyingChairs/FlyingThings3D mixture and no Sintel data, runs for 4 epochs with a learning rate of $2\times10^{-4}$, and weights the 14 flow outputs with the same $\gamma=0.8$ as \cref{eq:loss}.
\section{Experiments}
\label{sec:exp}

\subsection{Experimental Setup}
\label{sec:exp_setup}
We train the reference iterative networks ourselves on FlyingChairs~\cite{dosovitskiy2015flownet} only, with the official settings of the FlyingChairs stage of each paper, because they serve to compare the linear output layer with \sem under identical training (\cref{sec:exp_general}).
Our final \ours is additionally fine-tuned on a mixture with FlyingThings3D~\cite{mayer2016things3d} (\cref{sec:training}), and EdgeFlowNet and NanoFlowNet use their released weights.
No network is trained or calibrated on MPI Sintel~\cite{butler2012sintel}, which serves as a novel test set.
We report the endpoint error (EPE) in pixels over 1041 pairs of Sintel images (both clean and final pass) and 640 FlyingChairs validation pairs, at the native resolution of each dataset.
INT4 denotes 4-bit weights with 8-bit activations.
The deployable networks, NibbleFlow, EdgeFlowNet and NanoFlowNet, are quantized with the Hailo Dataflow Compiler and its quantization-aware fine-tuning. 
We deploy on a Hailo-8 attached to a Raspberry~Pi~5, measure the frames per second (FPS) with a batch size of 1 on the Hailo-8 and on the CPU of the Raspberry~Pi~5, and count the GMACs per $448{\times}1024$ pair.
For the reference iterative networks, which cannot be deployed on the Hailo-8, and for the linear version of \ours, which replaces \sem by a linear output layer and is trained on FlyingChairs only, we simulate this quantization in PyTorch.
The weights of every convolution are rounded symmetrically to 4 bits with one scale per output channel, and every activation, including the correlation, the hidden state and the running flow, is quantized to 8 bits with its own range in each unrolled iteration.
The ranges are the minimum and maximum over 1024 FlyingChairs training pairs, and no additional fine-tuning follows.
Only in IRR-PWC are two layers kept in 8 bits in both versions, the second convolution of the first pyramid level and the first convolution of the flow estimator, since they are the two most sensitive to 4-bit weights.

\subsection{Comparison with Edge Optical Flow Networks}
\label{sec:exp_main}

\Cref{tab:main} compares \ours with the networks that can run on an integer edge accelerator.
In FP32, \ours is the most accurate of them by a wide margin, with a Sintel EPE of 3.88 and 4.76 on the clean and final pass against 7.41 and 8.25 for EdgeFlowNet, which has $5.4\times$ more parameters.
More importantly, this advantage survives quantization.
With 86\% of its weights in 4 bits, \ours loses less than one pixel of EPE and reaches 4.76 and 5.68, which cuts the error of EdgeFlowNet, the best deployable baseline, by up to 45\%.
NanoFlowNet is smaller and faster, but its INT4 error is more than twice as high.

The reference iterative networks, RAFT-small, RAPIDFlow-it12 and IRR-PWC, remain more accurate in FP32.
However, they sample their features at the current flow and therefore cannot be compiled.
They are nevertheless the natural test of our main claim, since they refine with a linear update that \sem can replace, which we examine in \cref{sec:exp_general}.

\subsection{Linear versus \sem Refinement}
\label{sec:exp_steps}
We first isolate the effect of the update representation.
To this end, we train the same network on FlyingChairs only with 6/6/3 iterations and the same seed, and change only the output of its update block, from the linear update to SEM.


\paragraph{Accuracy}
\sem has the lower Sintel EPE on both passes, both at the selected checkpoint and on average over the last 50 epochs of training (\cref{tab:linear_sem}).
Averaged over the last 50 epochs, it leads the linear update by 0.09 and 0.24 on the clean and final pass.

\paragraph{Dynamic range}
We then measure the quantity that \cref{sec:linear} predicts.
For every flow value, we compare the step that refinement needs with the step that the network takes, and we define the measured dynamic range as the ratio, in octaves, between the largest and the smallest median step taken.
At the first step of the $1/32$ level, \sem follows the needed step from $\pow{-6.9}$ to $\pow{0.1}$ pixels, a range of 7.0 octaves that matches the span of its exponents.
The linear update, in contrast, saturates below one pixel, with a largest step of $\pow{-0.6}$ pixels, and does not go below $\pow{-6.3}$ pixels, which leaves it 5.6 octaves (\cref{tab:linear_sem}).
\sem is therefore wider at both ends, and this holds at all three levels, as the analysis of \cref{sec:linear} predicts.


\begin{table}[t]
\centering
\caption{Linear and \sem updates in an otherwise identical network}
\label{tab:linear_sem}
\small
\setlength{\tabcolsep}{3pt}
\begin{tabular}{@{}l cc cc cc@{}}
\toprule
 & \multicolumn{2}{c}{Checkpoint} & \multicolumn{2}{c}{Last 50 epochs} & Range & Max \\
\cmidrule(lr){2-3}\cmidrule(lr){4-5}
Update & Clean & Final & Clean & Final & (oct.) & step \\
\midrule
Linear & 4.89 & 6.10 & 5.04{\scriptsize$\pm$.07} & 6.23{\scriptsize$\pm$.07} & 5.6 & $\pow{-0.6}$ \\
\sem{} & \textbf{4.85} & \textbf{5.84} & \textbf{4.95}{\scriptsize$\pm$.06} & \textbf{5.99}{\scriptsize$\pm$.06} & \textbf{7.0} & $\bm{\pow{0.1}}$ \\
\bottomrule
\end{tabular}
\end{table}


\setcounter{topnumber}{1}
\subsection{\sem Refinement in Existing Flow Networks}
\label{sec:exp_general}
\begin{figure*}[t]
  \centering
  \includegraphics[width=\linewidth]{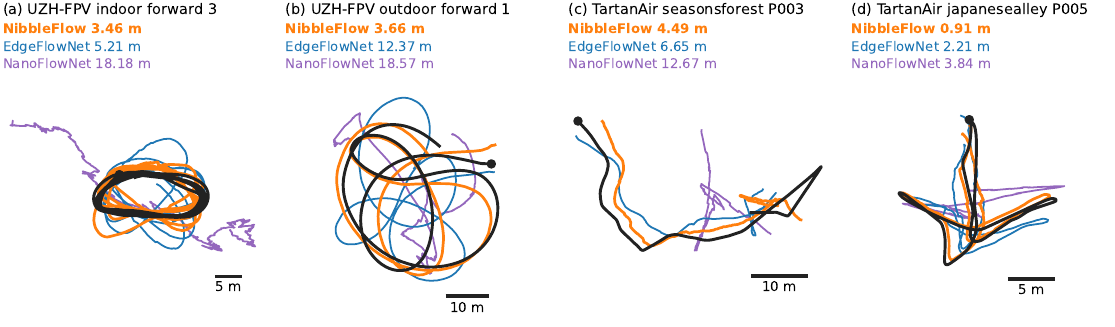}
  \caption{Visual odometry trajectories from optical flow alone with the INT4 models.
  Estimated paths of \ours (orange), EdgeFlowNet (blue) and NanoFlowNet (purple) against the ground truth (black) for two UZH-FPV flights (a, b) and two TartanAir trajectories (c, d), with each model's ATE on that sequence.
  The black dot marks the start of the trajectory.
  \ours has the lowest ATE on all four and keeps the shape of the loops (a, b) and of the turns (c, d); EdgeFlowNet keeps the shape but drifts further from the ground truth path, and the path of NanoFlowNet jitters and leaves it.}
  \label{fig:vo_traj_main}
\end{figure*}
If the linear output is indeed the limit, then replacing it should help other iterative networks as well.
We test this on networks that refine very differently from \ours.
RAFT-small~\cite{teed2020raft} iterates a GRU at a single scale, RAPIDFlow~\cite{morimitsu2024rapidflow} refines from coarse to fine with a shared block, and IRR-PWC~\cite{hur2019irr} refines a feature pyramid with shared weights.
In each network, we replace the convolution that regresses the update by the \sem head of \cref{eq:em}, with the same exponents for all networks and no tuning, and change nothing else.
Both versions are trained from scratch on FlyingChairs with the official settings and the same seed.

\paragraph{Floating point}
Replacing the output layer costs no accuracy in RAFT-small and RAPIDFlow.
At the official iteration counts, the two versions differ by less than the variation between checkpoints, with 2.80 against 2.84 on Sintel clean for RAFT-small and 2.57 for both versions of RAPIDFlow.
With a single iteration, \sem is ahead in both networks, by 0.40 EPE in RAFT-small.

\paragraph{4-bit weights}
The picture changes completely once the weights are quantized to 4 bits (\cref{tab:main}).
We quantize both versions of each network identically, with 4-bit weights and 8-bit activations.
The linear RAPIDFlow then loses 2.28 EPE on Sintel clean at its 12 iterations and 4.41 with a single iteration, whereas the \sem version, at 2.57 and 5.23 in FP32, loses only 0.70 and 0.61.
In other words, \sem removes up to 86\% of the error that 4-bit weights add.
The effect is the same in RAFT-small, where the error of the linear version rises from 2.84 to 17.83 and that of \sem only to 8.79, and in IRR-PWC.
\Cref{fig:ptq_qual} shows the same effect on individual examples.
With the linear output layer, the INT4 flow is washed out, noisy or dominated by a spurious motion, whereas with \sem it stays close to the FP32 result in all four networks.



\paragraph{Attribution of the quantization error}
Quantizing one part of the network at a time locates this loss in a single layer.
In RAPIDFlow, the 4-bit weights of the last convolution alone, which is the layer that \sem replaces, account for nearly all of the loss of the linear version.
With a single iteration, they cost the linear version 4.98 of the 5.00 EPE that quantizing every layer costs in this diagnostic, and \sem only 0.10, while the 8-bit grid of the step costs either version at most 0.07.
This is the error of \cref{sec:linear} in its plainest form.
The rounding of the output weights passes unchanged into every linear step and accumulates over the iterations, whereas in \sem it perturbs a sign, an exponent and a mantissa, whose effect is relative to the step.
Keeping this one layer in 8 bits also rescues the linear version, which then matches \sem.
The value of \sem is that its output layer tolerates 4-bit weights like every other layer, so no layer needs special treatment.


\subsection{Visual Odometry Evaluation}
\label{sec:exp_vo}
\begin{table}[t]
\centering
\caption{Visual odometry from optical flow alone with the INT4 models
}
\label{tab:vo_main}
\footnotesize
\setlength{\tabcolsep}{2pt}
\begin{tabular}{@{}ccc ccc ccc c@{}}
\toprule
\multicolumn{3}{c}{\textbf{\ours (ours)}} & \multicolumn{3}{c}{EdgeFlowNet} & \multicolumn{3}{c}{NanoFlowNet} & GT \\
\cmidrule(r){1-3}\cmidrule(lr){4-6}\cmidrule(lr){7-9}\cmidrule(l){10-10}
ATE$\downarrow$ & Rot.$\downarrow$ & Dir.$\downarrow$ & ATE$\downarrow$ & Rot.$\downarrow$ & Dir.$\downarrow$ & ATE$\downarrow$ & Rot.$\downarrow$ & Dir.$\downarrow$ & ATE \\
\midrule
\multicolumn{10}{c}{\emph{TartanAir, synthetic}} \\
\midrule
\textbf{4.95} & 0.37 & \textbf{31.84} & 6.70 & \textbf{0.36} & 33.41 & 10.64 & 1.27 & 74.73 & 0.26 \\
\midrule
\multicolumn{10}{c}{\emph{UZH-FPV, real drone flights}} \\
\midrule
\textbf{5.46} & \textbf{0.34} & 12.48 & 8.89 & 0.39 & \textbf{12.01} & 18.76 & 1.84 & 71.09 & -- \\
\bottomrule
\end{tabular}
\tabnotes{Unweighted means over the sequences. Bold marks the best of the three networks, and GT is the ground truth flow.}
\end{table}

Dense flow on a small robot is rarely the end product, so we test whether the ranking of \cref{tab:main} survives a downstream task.
\Cref{tab:vo_main} reports monocular visual odometry (VO) from optical flow alone on eight synthetic TartanAir trajectories~\cite{tartanair} and three real UZH-FPV drone flights~\cite{uzhfpv}.
The three deployable networks run in INT4 and feed the same pose estimator, the five point algorithm inside RANSAC.
No depth is used and the translation scale is taken from the ground truth, so the metrics measure only how well the flow constrains the camera rotation and the direction of travel.
We report the ATE, the RMSE of the camera position in meters after SE(3) alignment, together with the median rotation error per frame (Rot.)\ and the mean error of the translation direction per frame (Dir.), both in degrees.
On TartanAir, \ours has the lowest ATE on five of the eight trajectories and the lowest mean ATE, 4.95\,m against 6.70\,m for EdgeFlowNet, with the largest margins on japanesealley P005, neighborhood P002 and amusement P007, where its ATE is 59\%, 54\% and 45\% lower.
The two networks are close in rotation, with a mean error of $0.37^\circ$ and $0.36^\circ$, and \ours has the lower mean direction error, $31.8^\circ$ against $33.4^\circ$.
NanoFlowNet drifts more than twice as far as \ours, with a mean ATE of 10.64\,m and more than twice its direction error.
The ground truth flow through the same estimator gives a mean ATE of 0.26\,m, an order of magnitude below every network, so the remaining error lies in the flow rather than in the estimator.
UZH-FPV provides no ground truth flow.
On the real flights, \ours has the lowest ATE on indoor\_forward\_3 and outdoor\_forward\_1, with 3.46\,m and 3.66\,m against 5.21\,m and 12.37\,m for EdgeFlowNet, and the two are within 0.2\,m on indoor\_45\_2.
Its mean ATE is therefore 5.46\,m against 8.89\,m, and its mean rotation error is also the lowest, whereas EdgeFlowNet has a slightly lower mean direction error, $12.0^\circ$ against $12.5^\circ$.
Over all eleven sequences, \ours has the lowest ATE on seven, EdgeFlowNet on four and NanoFlowNet on none.

\Cref{fig:vo_traj_main} overlays the estimated paths on the ground truth for two flights and two trajectories.
\ours follows the repeated loops of the indoor flight (a) and the large loops of the outdoor flight (b), where EdgeFlowNet keeps the shape but drifts further from the ground truth path, and it stays closer to the turns of seasonsforest (c) and japanesealley (d).
The path of NanoFlowNet jitters and leaves the ground truth in every panel.
The VO results thus confirm the EPE ranking of \cref{tab:main}, since among the deployable networks \ours produces the flow most consistent with the camera motion, with $5.4\times$ fewer parameters than EdgeFlowNet.
\section{Conclusion}
\label{sec:conclusion}

We presented \textbf{\textcolor{semgreen}{SEM}}, a refinement step that represents each flow update as a \textbf{\textcolor{semgreen}{S}}ign, an \textbf{\textcolor{semgreen}{E}}xponent and a \textbf{\textcolor{semgreen}{M}}antissa.
Our analysis shows that a linear update has a fixed dynamic range, which is precisely what breaks iterative optical flow once its weights are quantized to 4 bits.
\sem removes this limit.
It covers the full range of steps that refinement needs and makes the output layer as tolerant of 4-bit weights as any other layer.
Without any tuning, it removes up to 86\% of the error that 4-bit weights add to existing iterative networks.
Built around \sem, \ours runs on a Hailo-8 with only 0.51M parameters and most of its weights in 4 bits, and it is the most accurate deployable network in our comparison.
We believe that the same principle applies to any iterative regressor that must run in fixed point arithmetic, such as depth or disparity refinement.
We will release our code and trained checkpoints upon acceptance.


{
    \small
    \bibliographystyle{IEEEtran}
    \bibliography{references}

@article{chao2014survey,
  title   = {A Survey of Optical Flow Techniques for Robotics Navigation Applications},
  author  = {Chao, Haiyang and Gu, Yu and Napolitano, Marcello},
  journal = {Journal of Intelligent \& Robotic Systems},
  volume  = {73},
  number  = {1},
  pages   = {361--372},
  year    = {2014}
}

@article{pandey2021leveraging,
  title   = {Leveraging Deep Learning for Visual Odometry Using Optical Flow},
  author  = {Pandey, Tejas and Pena, Dexmont and Byrne, Jonathan and Moloney, David},
  journal = {Sensors},
  volume  = {21},
  number  = {4},
  pages   = {1313},
  year    = {2021}
}

@inproceedings{schonberger2016structure,
  title     = {Structure-from-Motion Revisited},
  author    = {Sch{\"o}nberger, Johannes L. and Frahm, Jan-Michael},
  booktitle = {IEEE Conference on Computer Vision and Pattern Recognition (CVPR)},
  pages     = {4104--4113},
  year      = {2016}
}

@article{karpuzov2024object,
  title   = {Object Tracking Based on Optical Flow Reconstruction of Motion-Group Parameters},
  author  = {Karpuzov, Simeon and Petkov, George and Ilieva, Sylvia and Petkov, Alexander and Kalitzin, Stiliyan},
  journal = {Information},
  volume  = {15},
  number  = {6},
  pages   = {296},
  year    = {2024}
}

@inproceedings{bouwmeester2023nanoflownet,
  title     = {{NanoFlowNet}: Real-Time Dense Optical Flow on a Nano Quadcopter},
  author    = {Bouwmeester, Rik J. and Paredes-Vall{\'e}s, Federico and de Croon, Guido C. H. E.},
  booktitle = {IEEE International Conference on Robotics and Automation (ICRA)},
  pages     = {1996--2003},
  year      = {2023}
}

@article{edgeflownet2024,
  title   = {{EdgeFlowNet}: 100{FPS}@1{W} Dense Optical Flow for Tiny Mobile Robots},
  author  = {{Pinnama Raju}, Sai Ramana Kiran and Singh, Rishabh and Velmurugan, Manoj and Sanket, Nitin J.},
  journal = {IEEE Robotics and Automation Letters},
  volume  = {10},
  number  = {1},
  pages   = {128--135},
  year    = {2025}
}

@article{krispin2024real,
  title   = {Real-Time {3D} Object Detection Using {InnovizOne} {LiDAR} and Low-Power {Hailo-8} {AI} Accelerator},
  author  = {Krispin-Avraham, Itay and Orfaig, Roy and Bobrovsky, Ben-Zion},
  journal = {arXiv preprint arXiv:2412.05594},
  year    = {2024}
}

@inproceedings{dosovitskiy2015flownet,
  title     = {{FlowNet}: Learning Optical Flow with Convolutional Networks},
  author    = {Dosovitskiy, Alexey and Fischer, Philipp and Ilg, Eddy and H{\"a}usser, Philipp and Hazirbas, Caner and Golkov, Vladimir and van der Smagt, Patrick and Cremers, Daniel and Brox, Thomas},
  booktitle = {IEEE International Conference on Computer Vision (ICCV)},
  pages     = {2758--2766},
  year      = {2015}
}

@inproceedings{sun2018pwc,
  title     = {{PWC-Net}: {CNNs} for Optical Flow Using Pyramid, Warping, and Cost Volume},
  author    = {Sun, Deqing and Yang, Xiaodong and Liu, Ming-Yu and Kautz, Jan},
  booktitle = {IEEE/CVF Conference on Computer Vision and Pattern Recognition (CVPR)},
  pages     = {8934--8943},
  year      = {2018}
}

@inproceedings{teed2020raft,
  title     = {{RAFT}: Recurrent All-Pairs Field Transforms for Optical Flow},
  author    = {Teed, Zachary and Deng, Jia},
  booktitle = {European Conference on Computer Vision (ECCV)},
  pages     = {402--419},
  year      = {2020}
}

@inproceedings{kong2021fastflownet,
  title     = {{FastFlowNet}: A Lightweight Network for Fast Optical Flow Estimation},
  author    = {Kong, Lingtong and Shen, Chunhua and Yang, Jie},
  booktitle = {IEEE International Conference on Robotics and Automation (ICRA)},
  pages     = {10310--10316},
  year      = {2021}
}

@inproceedings{zhang2025neuflow,
  title     = {{NeuFlow-V2}: Push High-Efficiency Optical Flow to the Limit},
  author    = {Zhang, Zhiyong and Gupta, Aniket and Jiang, Huaizu and Singh, Hanumant},
  booktitle = {IEEE/RSJ International Conference on Intelligent Robots and Systems (IROS)},
  pages     = {2479--2485},
  year      = {2025}
}

@inproceedings{jacob2018quantization,
  title     = {Quantization and Training of Neural Networks for Efficient Integer-Arithmetic-Only Inference},
  author    = {Jacob, Benoit and Kligys, Skirmantas and Chen, Bo and Zhu, Menglong and Tang, Matthew and Howard, Andrew and Adam, Hartwig and Kalenichenko, Dmitry},
  booktitle = {IEEE Conference on Computer Vision and Pattern Recognition (CVPR)},
  pages     = {2704--2713},
  year      = {2018}
}

@inproceedings{mayer2016things3d,
  title     = {A Large Dataset to Train Convolutional Networks for Disparity, Optical Flow, and Scene Flow Estimation},
  author    = {Mayer, Nikolaus and Ilg, Eddy and H{\"a}usser, Philipp and Fischer, Philipp and Cremers, Daniel and Dosovitskiy, Alexey and Brox, Thomas},
  booktitle = {IEEE Conference on Computer Vision and Pattern Recognition (CVPR)},
  pages     = {4040--4048},
  year      = {2016}
}

@inproceedings{tartanair,
  title     = {{TartanAir}: A Dataset to Push the Limits of Visual {SLAM}},
  author    = {Wang, Wenshan and Zhu, Delong and Wang, Xiangwei and Hu, Yaoyu and Qiu, Yuheng and Wang, Chen and Hu, Yafei and Kapoor, Ashish and Scherer, Sebastian},
  booktitle = {IEEE/RSJ International Conference on Intelligent Robots and Systems (IROS)},
  pages     = {4909--4916},
  year      = {2020}
}

@inproceedings{uzhfpv,
  title     = {Are We Ready for Autonomous Drone Racing? {The} {UZH-FPV} Drone Racing Dataset},
  author    = {Delmerico, Jeffrey and Cieslewski, Titus and Rebecq, Henri and Faessler, Matthias and Scaramuzza, Davide},
  booktitle = {IEEE International Conference on Robotics and Automation (ICRA)},
  pages     = {6713--6719},
  year      = {2019}
}

@inproceedings{fu2018dorn,
  title     = {Deep Ordinal Regression Network for Monocular Depth Estimation},
  author    = {Fu, Huan and Gong, Mingming and Wang, Chaohui and Batmanghelich, Kayhan and Tao, Dacheng},
  booktitle = {IEEE/CVF Conference on Computer Vision and Pattern Recognition (CVPR)},
  pages     = {2002--2011},
  year      = {2018}
}

@inproceedings{bhat2021adabins,
  title     = {{AdaBins}: Depth Estimation Using Adaptive Bins},
  author    = {Bhat, Shariq Farooq and Alhashim, Ibraheem and Wonka, Peter},
  booktitle = {IEEE/CVF Conference on Computer Vision and Pattern Recognition (CVPR)},
  year      = {2021}
}

@inproceedings{kendall2017gcnet,
  title     = {End-to-End Learning of Geometry and Context for Deep Stereo Regression},
  author    = {Kendall, Alex and Martirosyan, Hayk and Dasgupta, Saumitro and Henry, Peter and Kennedy, Ryan and Bachrach, Abraham and Bry, Adam},
  booktitle = {IEEE International Conference on Computer Vision (ICCV)},
  pages     = {66--75},
  year      = {2017}
}

@inproceedings{xu2022gmflow,
  title     = {{GMFlow}: Learning Optical Flow via Global Matching},
  author    = {Xu, Haofei and Zhang, Jing and Cai, Jianfei and Rezatofighi, Hamid and Tao, Dacheng},
  booktitle = {IEEE/CVF Conference on Computer Vision and Pattern Recognition (CVPR)},
  pages     = {8121--8130},
  year      = {2022}
}

@inproceedings{li2020gfl,
  title     = {Generalized Focal Loss: Learning Qualified and Distributed Bounding Boxes for Dense Object Detection},
  author    = {Li, Xiang and Wang, Wenhai and Wu, Lijun and Chen, Shuo and Hu, Xiaolin and Li, Jun and Tang, Jinhui and Yang, Jian},
  booktitle = {Advances in Neural Information Processing Systems (NeurIPS)},
  year      = {2020}
}

@inproceedings{farebrother2024stop,
  title     = {Stop Regressing: Training Value Functions via Classification for Scalable Deep {RL}},
  author    = {Farebrother, Jesse and Orbay, Jordi and Vuong, Quan and Ta{\"i}ga, Adrien Ali and Chebotar, Yevgen and Xiao, Ted and Irpan, Alex and Levine, Sergey and Castro, Pablo Samuel and Faust, Aleksandra and Kumar, Aviral and Agarwal, Rishabh},
  booktitle = {International Conference on Machine Learning (ICML)},
  series    = {PMLR},
  volume    = {235},
  pages     = {13049--13071},
  year      = {2024}
}

@article{lee2019gsd,
  title   = {Regression Tree {CNN} for Estimation of Ground Sampling Distance Based on Floating-Point Representation},
  author  = {Lee, Jae-Hun and Sull, Sanghoon},
  journal = {Remote Sensing},
  volume  = {11},
  number  = {19},
  pages   = {2276},
  year    = {2019}
}

@article{charton2022linear,
  title   = {Linear Algebra with Transformers},
  author  = {Charton, Fran{\c{c}}ois},
  journal = {Transactions on Machine Learning Research (TMLR)},
  year    = {2022}
}

@article{hafner2025dreamerv3,
  title   = {Mastering Diverse Control Tasks through World Models},
  author  = {Hafner, Danijar and Pasukonis, Jurgis and Ba, Jimmy and Lillicrap, Timothy},
  journal = {Nature},
  volume  = {640},
  number  = {8059},
  pages   = {647--653},
  year    = {2025}
}

@article{miyashita2016logarithmic,
  title   = {Convolutional Neural Networks using Logarithmic Data Representation},
  author  = {Miyashita, Daisuke and Lee, Edward H. and Murmann, Boris},
  journal = {arXiv preprint arXiv:1603.01025},
  year    = {2016}
}

@article{micikevicius2022fp8,
  title   = {{FP8} Formats for Deep Learning},
  author  = {Micikevicius, Paulius and Stosic, Dusan and Burgess, Neil and Cornea, Marius and Dubey, Pradeep and Grisenthwaite, Richard and Ha, Sangwon and Heinecke, Alexander and Judd, Patrick and Kamalu, John and Mellempudi, Naveen and Oberman, Stuart and Shoeybi, Mohammad and Siu, Michael and Wu, Hao},
  journal = {arXiv preprint arXiv:2209.05433},
  year    = {2022}
}

@article{song2026tinyflow,
  title   = {{TinyFlow}: Breaking the Warping Bottleneck for Edge Optical Flow},
  author  = {Song, Xiao'ao and Singh, Chahat Deep},
  journal = {arXiv preprint arXiv:2610.33338},
  year    = {2026}
}

@inproceedings{wang2024searaft,
  title     = {{SEA-RAFT}: Simple, Efficient, Accurate {RAFT} for Optical Flow},
  author    = {Wang, Yihan and Lipson, Lahav and Deng, Jia},
  booktitle = {European Conference on Computer Vision (ECCV)},
  year      = {2024}
}

@inproceedings{hui2018liteflownet,
  title     = {{LiteFlowNet}: A Lightweight Convolutional Neural Network for Optical Flow Estimation},
  author    = {Hui, Tak-Wai and Tang, Xiaoou and Loy, Chen Change},
  booktitle = {IEEE/CVF Conference on Computer Vision and Pattern Recognition (CVPR)},
  year      = {2018}
}

@inproceedings{butler2012sintel,
  title     = {A Naturalistic Open Source Movie for Optical Flow Evaluation},
  author    = {Butler, Daniel J. and Wulff, Jonas and Stanley, Garrett B. and Black, Michael J.},
  booktitle = {European Conference on Computer Vision (ECCV)},
  year      = {2012}
}

@article{nagel2021white,
  title   = {A White Paper on Neural Network Quantization},
  author  = {Nagel, Markus and Fournarakis, Marios and Amjad, Rana Ali and Bondarenko, Yelysei and van Baalen, Mart and Blankevoort, Tijmen},
  journal = {arXiv preprint arXiv:2106.08295},
  year    = {2021}
}

@inproceedings{hur2019irr,
  title     = {Iterative Residual Refinement for Joint Optical Flow and Occlusion Estimation},
  author    = {Hur, Junhwa and Roth, Stefan},
  booktitle = {IEEE/CVF Conference on Computer Vision and Pattern Recognition (CVPR)},
  year      = {2019}
}

@inproceedings{morimitsu2024rapidflow,
  title     = {{RAPIDFlow}: Recurrent Adaptable Pyramids with Iterative Decoding for Efficient Optical Flow Estimation},
  author    = {Morimitsu, Henrique and Zhu, Xiaobin and Cesar-Jr., Roberto M. and Ji, Xiangyang and Yin, Xu-Cheng},
  booktitle = {IEEE International Conference on Robotics and Automation (ICRA)},
  year      = {2024}
}
}

\ifsupp
\clearpage
\input{supp}
\fi

\end{document}